# Crossbreeding in Random Forest

Abolfazl Nadi[a], Hadi Moradi[a,b]*, Khalil Taheri[a]

[a] Advanced Robotics and Intelligent Systems Laboratory, School of Electrical and Computer Engineering, College of Engineering, University of Tehran, Tehran, 11155-4563, Iran,
a.nadi@ut.ac.ir, moradih@ut.ac.ir, k.taheri@ut.ac.ir

[b] Intelligent System Research Institute (ISRI), SKKU, Suwon, 16419, South Korea,
moradih@ut.ac.ir

*(Corresponding author: moradih@ut.ac.ir, +98 21 61114960)

**Abstract**

Ensemble learning methods are designed to benefit from multiple learning algorithms for better predictive performance. The tradeoff of this improved performance is slower speed and larger size of ensemble learning systems compared to single learning systems. In this paper, we present a novel approach to deal with this problem in Random Forest (RF) as one of the most powerful ensemble methods. The method is based on crossbreeding of the best tree branches to increase the performance of RF in space and speed while keeping the performance in the classification measures. The proposed approach has been tested on a group of synthetic and real datasets and compared to the standard RF approach. Several evaluations have been conducted to determine the effects of the Crossbred RF (CRF) on the accuracy and the number of trees in a forest. The results show better performance of CRF compared to RF.

**Keywords:** pruning method, Random forest, Cross breeding, High-dimensional data, Classification.

# 1. Introduction

One of the most powerful and popular ensemble methods is RF that has a wide range of applications in data mining and machine learning, especially for classification of large high dimensional data (Brakenhoff& van de Wiel, 2017).

The following benefits can be recognized for RF in comparison to other classification algorithms (Xu et al., 2012): (1) As an ensemble method, in which each part of classifier is built from a subspace of data, it is proficient in modeling classes in subspaces. (2) Large datasets can be operated efficiently because of the use of decision tree generalization to build the component classifiers. (3) High dimensional data is well-operated for multi-class tasks, such as classifying text data, which have many varieties. (4) The component classifiers, within the ensemble, can be created in parallel in a distributed environment, significantly decreasing the time for creating an RF model from large datasets (Xu et al., 2012). Experimental results have shown that RF models can obtain high accuracy in data classification (Braga, Madureira, Coelho & Ajith, 2019), regression (Ceh, Kilibarda, Lisec & Bajat, 2018), and novelty detection (Cichosz et al., 2016). In comparison to the state-of-the-art supervised classification algorithms, the RF method gives higher accuracy (such as AdaBoost and SVM (Cano et al., 2017)). It has been shown that the RF method is simple, quick, easily parallelized, resistant to outliers and noise (Breiman, 2001), and does not overfit unlike Ada-Boost algorithm, which is unstable to noisy data (Breiman, 2001; Dietterich, 1998). Breiman (2001) has demonstrated that RF gives useful internal estimates of error, strength, correlation, and variable importance.

Based on these advantages, we will consider RF as our base algorithm to apply our crossbreeding approach. In an RF, each tree is constructed based on a selected group of features (subset of features). This is the advantage of RF over basic C4.5 decision tree and many other classification approaches, even other ensemble methods, since it uses many classifiers to make a decision to avoid the C4.5 local minima pitfall. In other words, the more trees are incorporated in an RF, the higher the accuracy of the classification and the lower the chance of getting stuck in local minima. Unfortunately, it is not possible since the number of trees is limited due to the size of each tree, which is determined by the number of informative features in a given problem (Fawagreh, Gaber & Elyan, 2014). For instance, in RF applications studied by Doumanoglou, Kim, Zhao and Malassiotis (2014), Fina, Karaman, Bagdanov and Bimbo (2015), Zhu, Shao and Lin (2013) , and Li, Wu and Radke (2015), the full feature vectors of multimodal sensors are used to construct the trees of forests. There are recent studies that address the size of trees in RF (Duroux and Scornet, 2016; Al-Janabi & Andras, 2017). In Duroux and Scornet (2016), the number of terminal nodes is controlled, and its performance is compared to full size RF trees. They showed that with smaller number of terminal nodes, compared to the standard RF, the error can be lower than the standard RF. However, in this study, the authors did not address the increase in the number of trees, the effects of the number of trees when the number of informative features is very high, and the effect of the depth directly. Al-Janabi and Andras (2017) have done a study, on low dimensional data space, i.e. under 40 informative features, to show the effect of size and number of trees in RF. Due to the limited number of informative features in this their study, the authors concluded that the number of trees in RF does not have a significant effect on the accuracy. In other words, the study has not fully analyzed the problem. Nadi and Moradi (2019) showed that the depth of the trees can be limited to a small number without reducing the accuracy. On the other hand, they showed that the number of trees, with limited depth, can be increased to improve the accuracy of the classifier.

In this paper, we introduce a novel crossbreeding approach, called Crossbreed RF (CRF), to further reduce the size of an RF while maintaining the same accuracy or even improving the accuracy compared to a typical RF. The rest of the paper is organized as follows: related works are explained in the next section. In the third section, new approach is described. Experimental results are presented in the fourth section, and finally the conclusion of this study is described.

## 2. Related work

Pruning the unimportant learners is important to select the best learners and improve the overall performance of an ensemble learner. For instance, if there are many base learners (classifiers) in an ensemble, the computation overhead would increase proportionally. That is why it is important to apply pruning in an RF, as an ensemble learning approach. That is why there are several studies [] tried to reduce the number of trees in RF classifiers and find the optimal subset of them.

Based on our research, there is not any procedures to design these learning groups. Most of the researchers use the trial and error approach to design the learning groups. In this approach, in the first phase a larger set of candidate classifiers are generated. Then, in the selection phase, several subsets of these candidate classifiers are chosen to get the highest accuracy. For the generation phase, there are technics such as wrapping and growing which are used to manipulate the learning set in order to achieve more candidate classifiers. In the selection phase, in order to get the highest accuracy, the optimal subset is selected using exhaustive enumeration. To be more specific, all possible subsets are generated, which is equal to 2^N subsets in which N is the number of the sets in first phase.

Therefore, to be able to reduce the computation complexity, we must more effective methods than exhaustive enumeration, using a suitable selection criteria.

The selection criteria can be based on filtering or wrapping method. In the filtering approach, determines the initial evaluation the combination mechanism, i.e. the prediction mechanism of the group's subsets are not considered. On the other hand, the wrapper method uses the reevaluation and considers the prediction mechanism in the subsets of the learning group by adding or removing the classifiers.

As an example, Kohavy [] uses the wrapping method instead of filter method in order to select the subsets of classifiers in designing ensemble.

Several studies have been done regarding selection approach of the best classifiers in designing an ensemble. Heuristic methods choose the best strategy. In other words, these methods choose the N top classifiers which have the best accuracy where N is a constant number. The disadvantage of these methods is that they only choose the classifiers based on their accuracy and does not consider the diversification of the base classifiers. On the other side, since it is not needed to evaluate the different subsets of classification, heuristic methods reduce the computation complexity.

Studies [] have shown that there are two principle approaches on generating diversified trees. The first approach tries to propose a basic and fundamental concept for diversification. The second approach use different technical concepts to generate diversified ensembles which based on two major methods, i.e. explicit and implicit methods.

Implicit methods use randomness to generate different paths in problem space. Explicit methods deterministically generate different paths into the space.

Despite of the technic that we use in order to diversify the trees of a forest, there are criteria for measuring the trees diversification in one specific method or between two different methods. These criteria are as follows: difference, duplicate error, KW variance, agreements between classifiers, generalized diversification, difficulty, Q statistics, and collaboration coefficient. Tang [] have reviewed and analyzed the mentioned criteria.

Search algorithms use a evaluation function for the evaluation of classifiers subsets in the selection section. Most of the time, both of the diversification and accuracy evaluation which is evaluated by a function is used as the evaluation criteria. These search algorithms are forward search and backward search. The forward search starts with a null subset. Classifiers are added to the subset one by one. In each step, the chosen classifier among the remaining classifiers increase the overall accuracy of the whole subset classifiers. On the other hand, the backward search starts with a full subsets of classifiers and, in each step, tries to remove the most useless classifier and at the same time gain the highest accuracy.

Methods based on the classification clustering tries to select the best classifiers as well. These methods have two phases. In the first phase, the clustering approach is used to get similar models and in the second phase, pruning of each cluster lead to the subset of classifiers which have the best possible accuracy and diversification. []

Recently, Giffon [] proposed a new approach in order to prune the random trees using the Orthogonal Matching Trees (OMT) approach. They introduce the concept of OMP forest and their experimental results shows the proposed method has excellent performance in term of ability to learn OMP weighs.

Based on the mentioned disadvantages of RF, in this paper, we have proposed CRF which tries to select the best tree branches to increase the accuracy of RF while reducing the size of RFs. This would have at least two benefits: a) the RF learning speed, per tree, is increased greatly, b) the size of trees stored in the memory is much smaller than standard RF trees.

## 3. The Proposed Method

To explain the proposed approach, the following notations would be used.

As previously mentioned, Schapire in 1990 [], and Breiman in years 1994 [], and 2001 [] defined a group forecaster f based on base learners called group methods. In classic random forest, each tree is defined as a base learner. As mentioned earlier, in the proposed approach, the branches are defined as the base learners. Therefore, the group forecaster f based on the base learners would be as follows:

$$\left(b_1(x), b_2(x), \ldots, b_j(x)\right) \tag{1}$$

Where j is the number of branches in the CRF. We will explain the modifications in classic random forest which would lead to our proposed approach.

Regression applications use average base learners as the follows:

$$f(X) = \frac{1}{J'} \sum_{j=1}^{J} \begin{cases} b_j(x) & if \quad b_j(x) \text{ is able to proceed } X \\ 0 & if \quad b_j(x) \text{ is NOT able to proceed } X \end{cases} \tag{1}$$

Where J′ is the number of base learners processed X.

In grouping, f(X) would be defined as follows:

$$f(X) = \begin{cases} \arg max_{y \epsilon \alpha} \sum_{j=1}^{J} I\left(y = b_j(x)\right) & \text{if } \in a\ b_j\ to\ proceed\ X \\ g(X) & otherwise \end{cases} \quad (2)$$

$$I(Y = f(x)) = \begin{cases} 1 & if\ Y\ is\ equal\ to\ f(X) \\ 0 & otherwise \end{cases} \quad (3)$$

$$g(X) = \begin{cases} A\ random\ decision \\ OR \\ A\ clustering\ based\ decision \end{cases} \quad (4)$$

where $g(X)$ is the decision function for those samples which are not under any branches.

In RFB method, the $J^{th}$ base learner is the branch which denotes by $b_j(x, \theta_j)$ where $\theta_j$ is the set of random variables. In classic RF, trees grow without any crossbreeding phase and they will grow until purification of the leaf nodes or the number of samples in leaf nodes lessees than the specified threshold. Random Forest of Branches (RFB) is based on branches of trees. Therefore, a new term is defined for trees' branches in order to divide trees to smaller units in random forest. Algorithm 1 shows the RFB approach based on the RF Algorithm in [].

The RFB algorithm uses the Binary Recursive Partitioning algorithm (BRP) to create trees in random forest.

Algorithm 1 - RFB

**Input:** $D = \{(x_1, y_1), \ldots, (x_N, y_N)\}$ that denotes the training data, with $x_i = (x_{i,1}, \ldots, x_{i,p})^T$

1. **For** $j \leftarrow 1$ to $J$
2. Take a bootstrap sample $D_j$ of size N from D.
3. Using the bootstrap sample $D_j$ as the training data, fit a tree using Binary Recursive Partitioning (BRP) (Algorithm 2):
4. Start with all observations at a single node.
5. Repeat the following steps recursively for each unsplit node:
6. Select $m$ predictors at random from the $p$ available predictors.
7. Find the best binary split among all binary splits on the $m$ predictors from Step (6).
8. Split the node into two descendant nodes using the split from Step (7).
9. **End of For**
10. Define **RFB** as an empty set of branches
11. **For** $j \leftarrow 1$ to $J$
12. **For** all $tn_i$ (terminal node (leaf nodes)) in $j^{th}$ tree

13. Define a $b_{i,j}$ as $i^{th}$ branch in $j^{th}$ tree by going from *root* to $tn_i$
14. Add $b_{i,j}$ to **RFB**
15. **End of For**
16. **End of For**
17. **RFB** ← using crossbreeding algorithm (Algorithm 3) for **RFB**
18. To make a prediction at a new point x, calculate $f(x)$ using equation (1) or (2).

**Output:** The resulted RFB

Algorithm 2- BRP

**Input:** $D = \{(x_1, y_1), ..., (x_N, y_N)\}$ that denotes the training data, with $x_i = (x_{i,1}, ..., x_{i,p})^T$

1. Start with all observations $(x_1, y_1), ..., (x_N, y_N)$ in a single node.
2. Repeat the following steps recursively for each unsplit node:
3. Find the best binary split among all binary splits on all $p$ predictors.
4. Split the node into two descendant nodes using the best split (Step 3).

**Output:** The resulted tree

Algorithm 3- Crossbreeding the random forest of trees based on the crossbreeding criteria

**Input:** $D = \{(x_1, y_1), ..., (x_N, y_N)\}$ that denotes the training data, with $b_i = (b_1, ..., b_J)$ that denotes the branches in RFB

1. Start with all observations $(x_1, y_1), ..., (x_N, y_N)$ in a single node.
2. Repeat the following steps recursively for each branch $b_i$:
3. Find $D_i$ as a subset of the training data can be reach to the terminal node of $b_i$
4. Find $acc_i$ the accuracy of $b_i$ over $D_i$ (Step 3)
5. Apply crossbreeding criteria and threshold to prun branches

**Output:** The resulted pruned RFB

Based on each crossbreeding criteria and their threshold, we would get different results which are explained in the next section.

## 4. Experimental Results

To evaluate the performance of our proposed method, we have implemented the approach and tested it on a synthetic dataset with the capability of parameter setting, like the number of samples, the number of features, the number of informative features, and the number of classes (Pedregosa et al., 2011). The synthetic dataset is used since real high dimensional datasets cannot be controlled to analyze all aspects of the proposed approach.

Furthermore, to show that the proposed approach works on real data, Epileptic Seizure Recognition Dataset [] from the UCI archive is used. It should be noted that the issues related to the noise are not considered in this research and experiments.

The Epileptic Seizure Recognition Dataset is a five-class dataset that is used to classify the Epileptic Seizure from the patients' EEG. In these five classes, the first class is labeled as Epileptic Seizure, and the other classess are labelled as non Epileptic Seizure. Therefore, without any effects on the experiments, the non Epileptic Seizure classes can be combined as a one new class. So that, the dataset converts ino a two-class dataset. Table 1 shows the specification on this dataset. It contains 179 integer-valued features. Fig. 1 demonstrates the statistical significance analysis of the features of the Epileptic Seizure Recognition dataset that includes mean and standard deviation for each feature.

**Table 1**
Specification of the Magic Gamma Telescope dataset.

| Number of instances | 11500 |
|---|---|
| Attribute characteristic | Integer |
| Number of features | 179 |
| Number of classes | 2 |

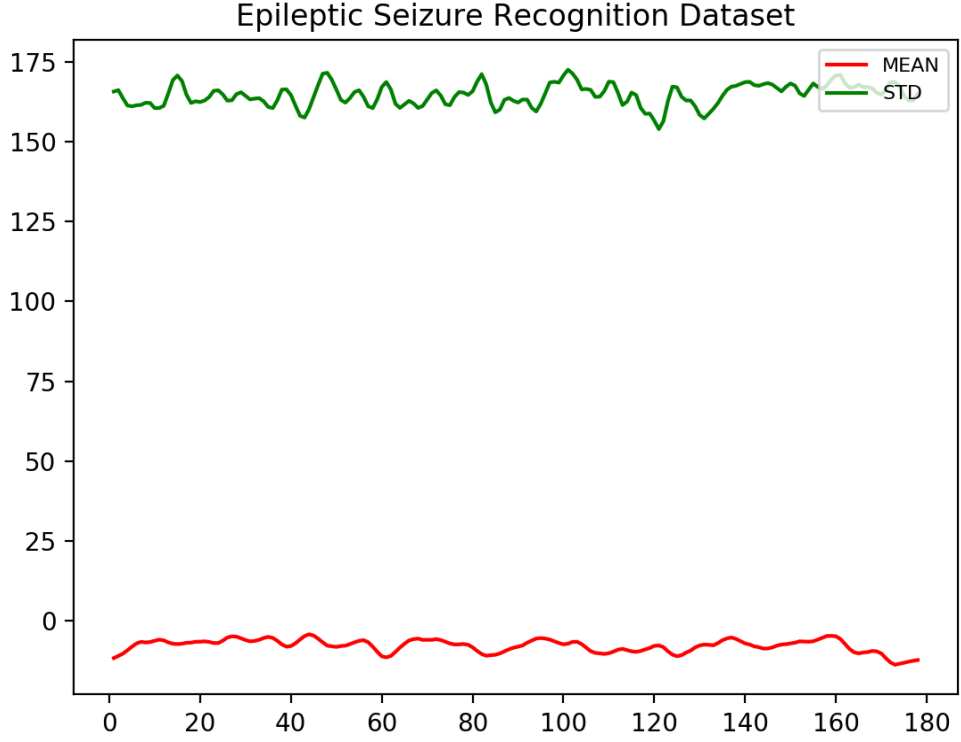

**Fig. 1.** Statistical significance analysis for the features of the Epileptic Seizure Recognition dataset.

In the following, first the analysis on the synthetic dataset is given followed by the analysis on the actual datasets.

*4.1. The accuracy criteria for existing branches in a RBF*

As previously mentioned, the accuracy of a branch in a random forest ($acc_{b_i}$) is calculated using the data which covered with the branch ($D_i$), and the leaf label which choosed for the covered data. This criteria only is used for classification applications. The accuracy of the $i^{th}$ branch is calculated as follows:

$$\forall x \in D_i : acc_{b_i} = \frac{1}{k_i} \sum_{j=1}^{k_i} I(l_i = label(x_j)) \quad (1)$$

$$I(Y = f(x)) = \begin{cases} 1 & \text{if } Y \text{ is equal to } f(X) \\ 0 & \text{otherwise} \end{cases} \quad (2)$$

Where $k_i$ is equal to the number of samples under coverage of the branch $b_i$ which is equal to $D_i$

## 4.2. The effect of the number of existing branches in trees on the accuracy

As previously mentioned, in order to show the effect of the proposed method on a random forest, it is needed to analysis the existing branches in trees at first. Therefore, the accuracy analysis of the existing branches in trees is evaluated first. Since, the train data are only available in order to prune the tress in a random forest, the following analysis are evaluated on the train data:

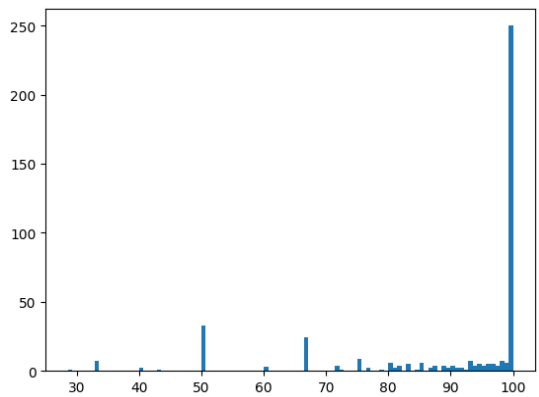

(a)

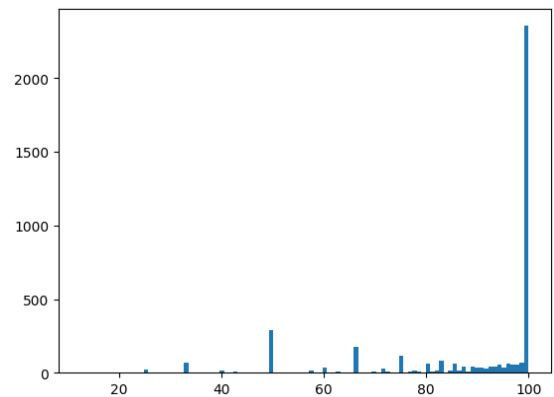

(b)

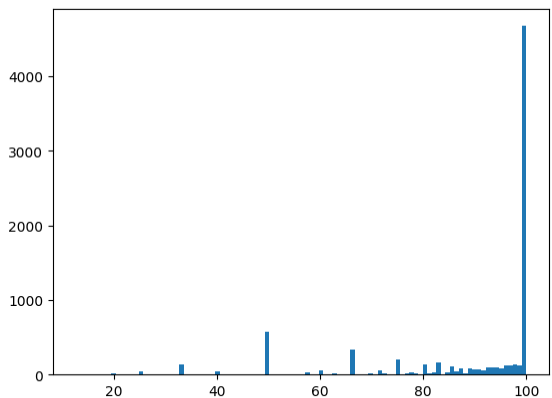

(c)

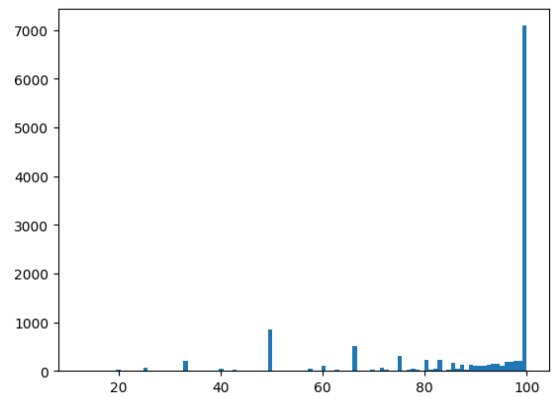

(d)

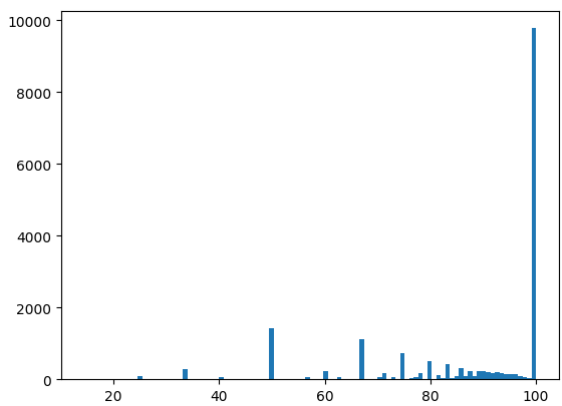

(e)

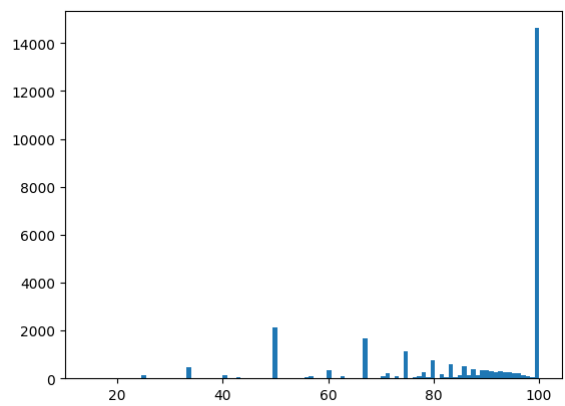

(f)

**Fig. 4.** The accuracy analysis of branches for data with one cluster in each feature for random forest with (a) one tree, (b) 10 trees, (c) 20 trees, (d) 30 trees, (e) 40 trees, (f) 50 trees.

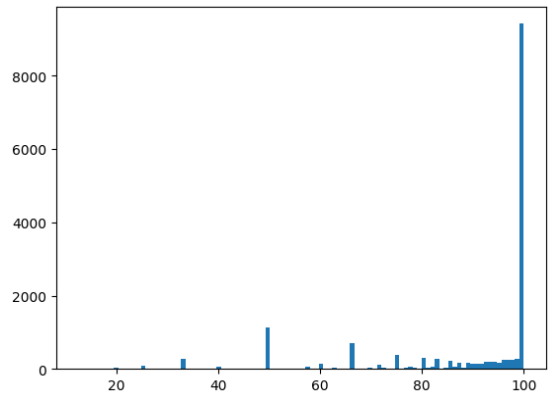

(a)

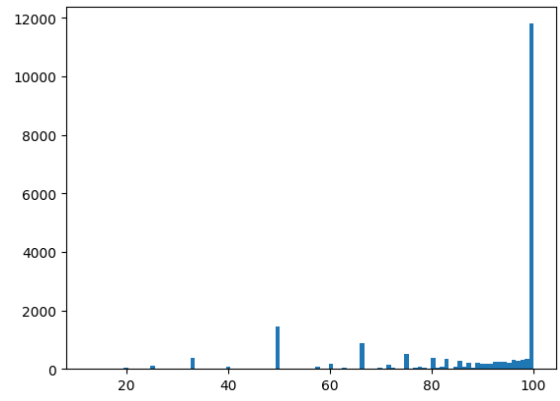

(b)

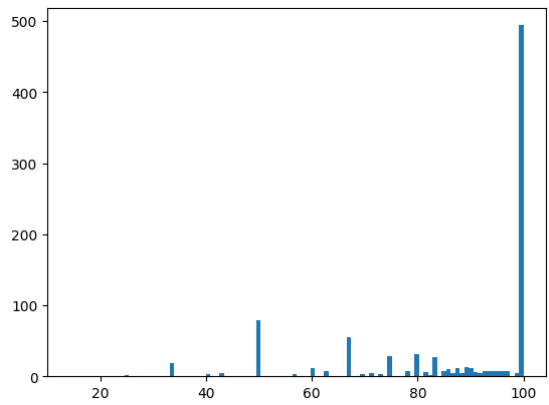

(c)

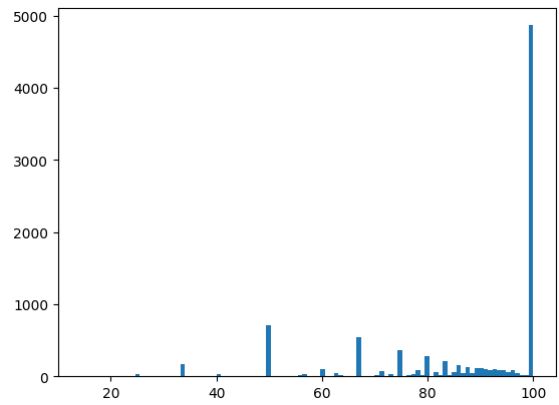

(d)

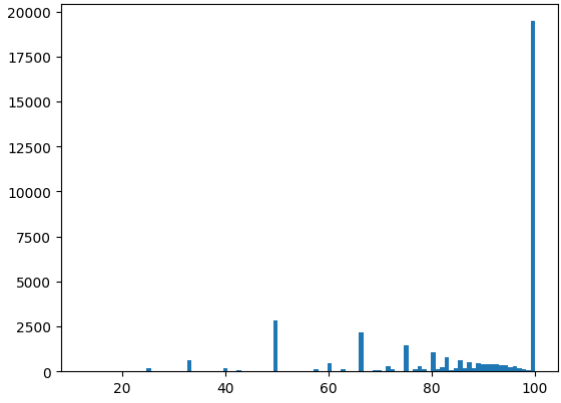
(e)

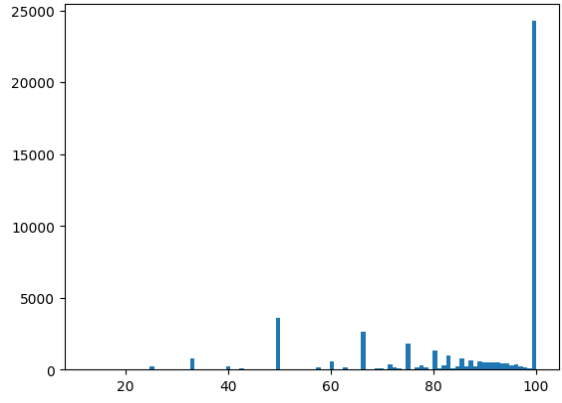
(f)

**Fig. 5.** The accuracy analysis of branches for data with 5 clusters in each feature for random forest with (a) one tree, (b) 10 trees, (c) 20 trees, (d) 30 trees, (e) 40 trees, (f) 50 trees.

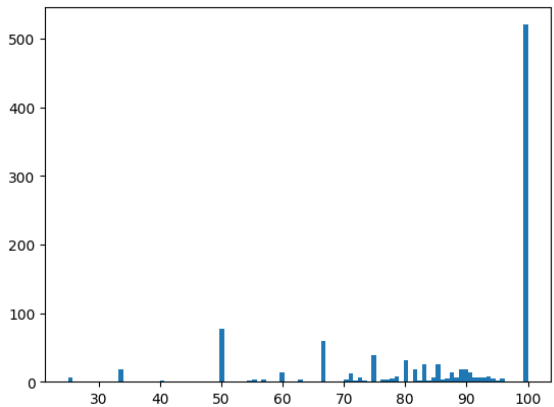
(a)

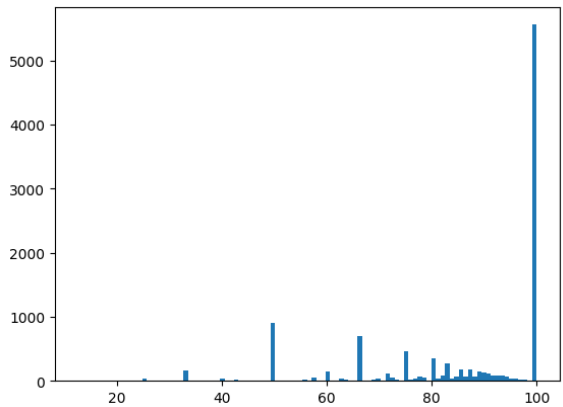
(b)

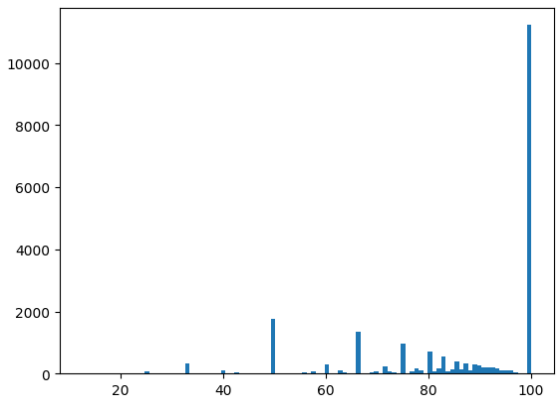
(c)

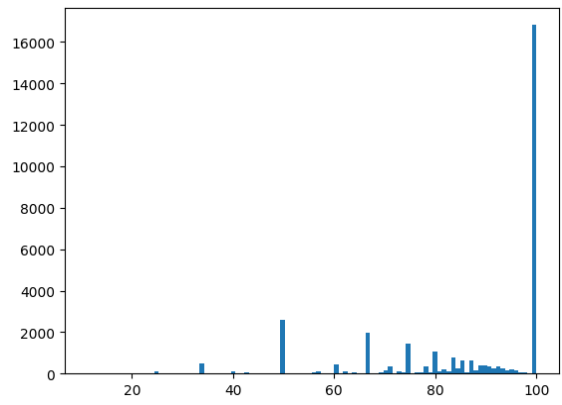
(d)

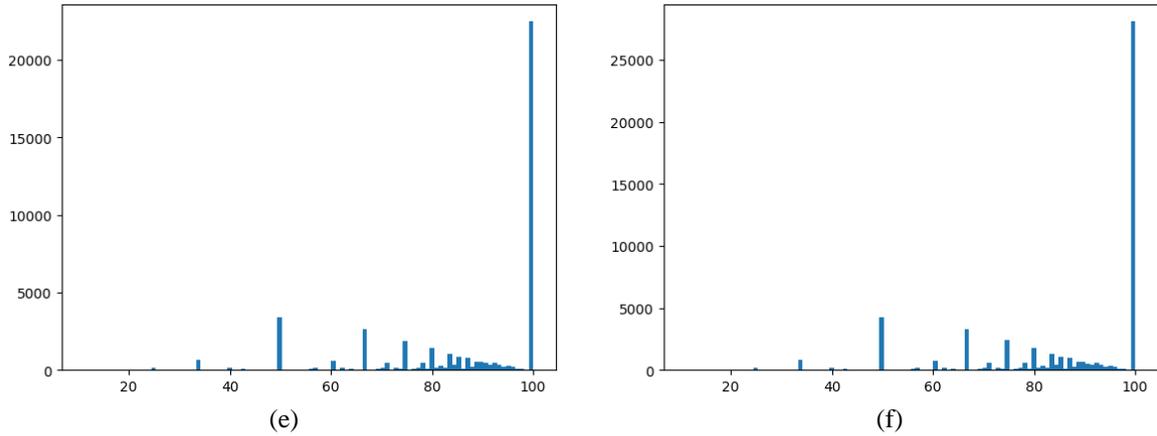

**Fig. 6.** The accuracy analysis of branches for data with 10 clusters in each feature for random forest with (a) one tree, (b) 10 trees, (c) 20 trees, (d) 30 trees, (e) 40 trees, (f) 50 trees.

It should be noted that, based on the above experiments, the considerable amount of existing branches in trees have the accuracy of 100%. Moreover, in all experiments, the data is split into the train and test data which are 75% and 25% of the whole data, respectively.

*4.3. The analysis of the number of existing branches in trees using the impact factor criteria*

As previously mentioned, in order to show the effect of the proposed method on a random forest, it is needed to analysis the existing branches in trees at first. Therefore, the accuracy analysis of the existing branches in trees is evaluated first. Since, the train data are only available in order to prune the tress in a random forest, the following analysis are evaluated on the train data:

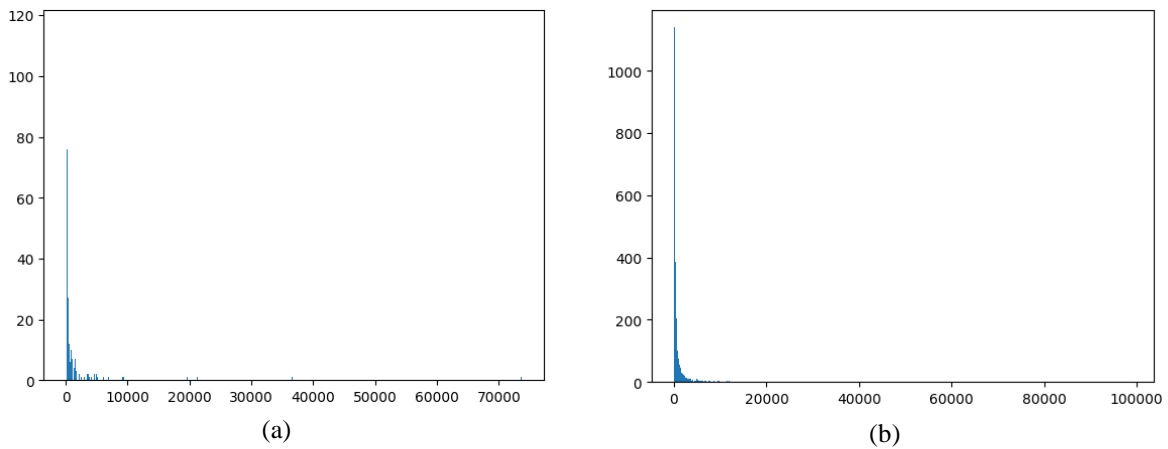

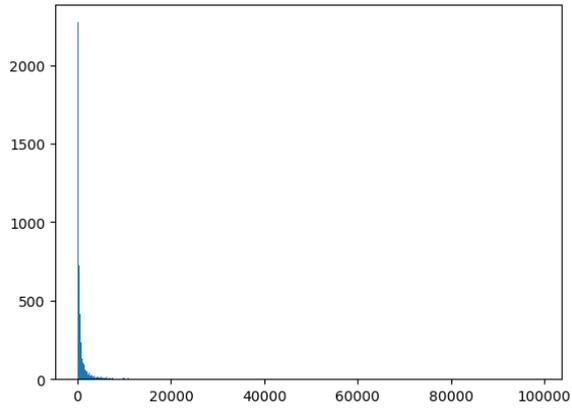
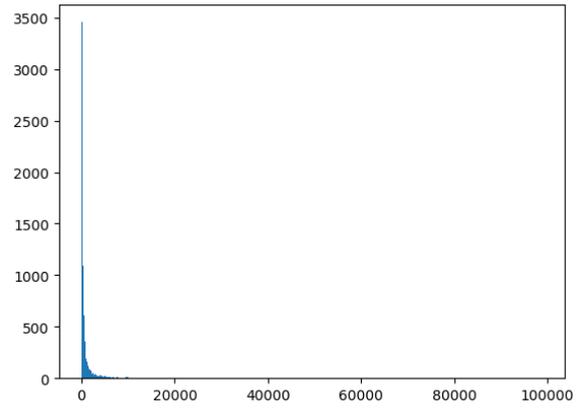
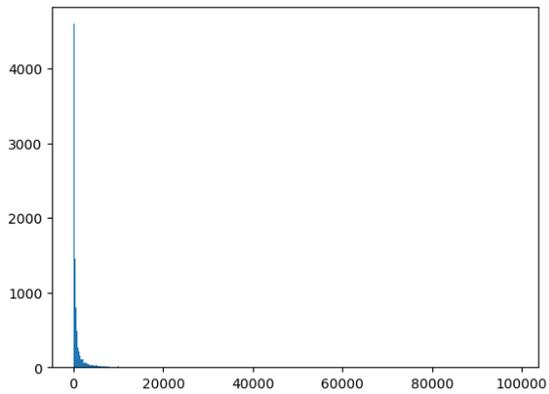
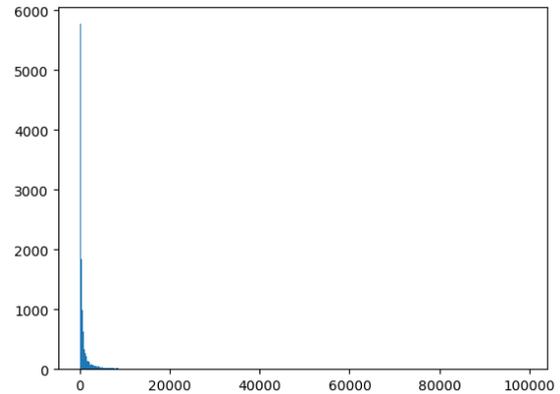
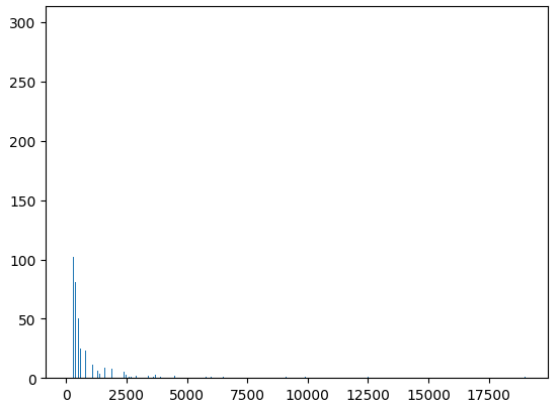
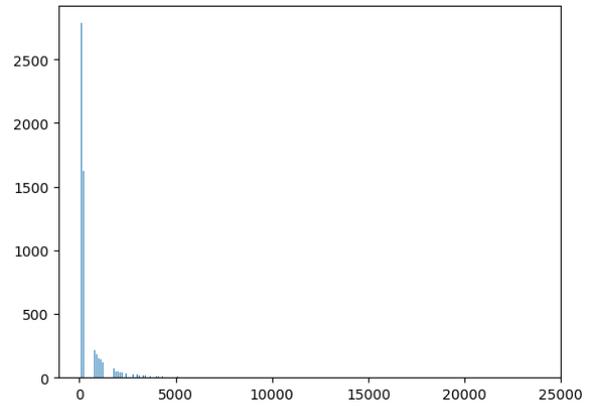

**Fig. 7**. The impact factor of branches for data with one cluster in each feature for random forest with (a) one tree, (b) 10 trees, (c) 20 trees, (d) 30 trees, (e) 40 trees, (f) 50 trees.

(a)

(b)

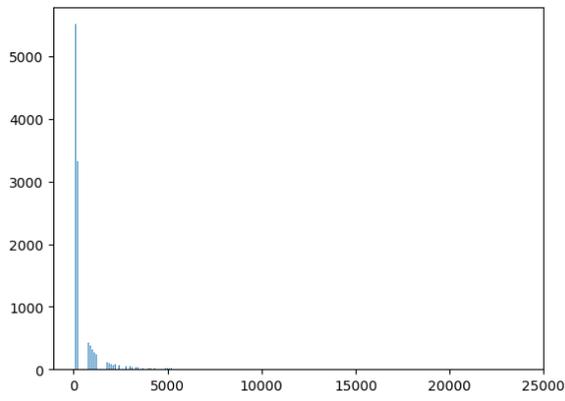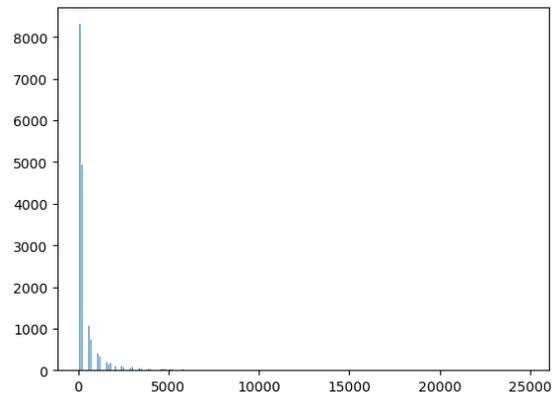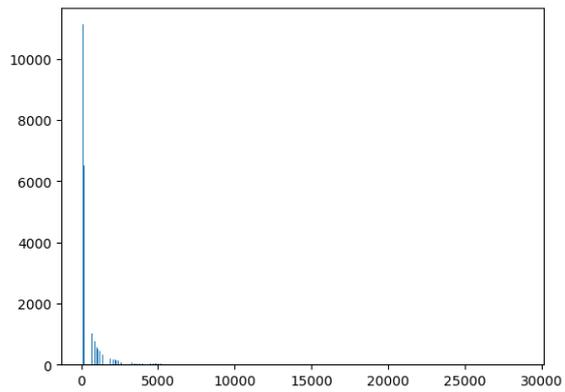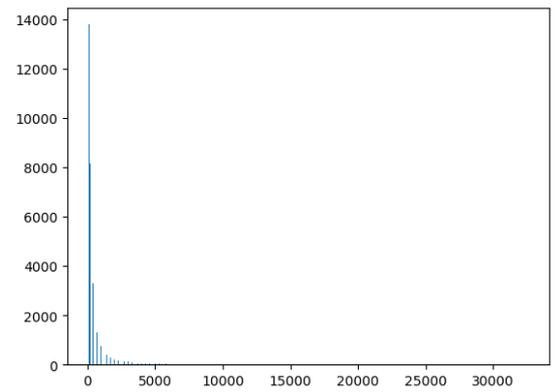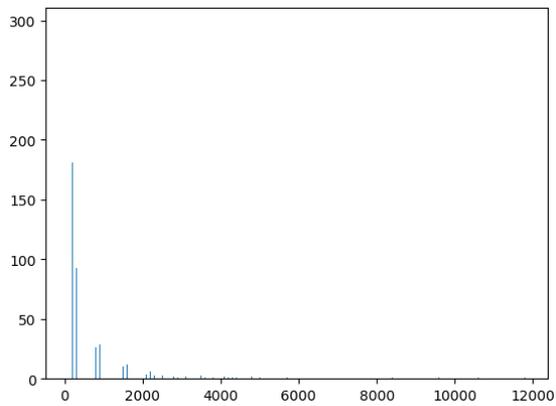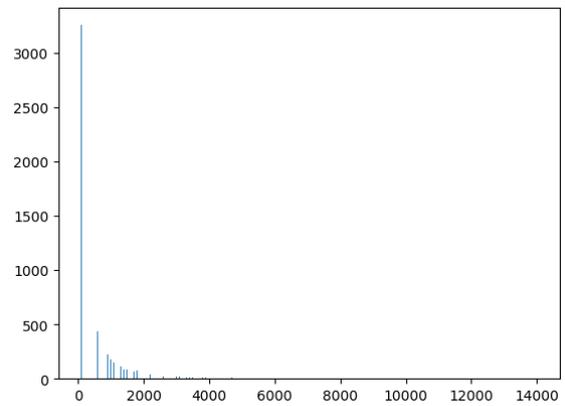

**Fig. 8**. The impact factor of branches for data with 5 clusters in each feature for random forest with (a) one tree, (b) 10 trees, (c) 20 trees, (d) 30 trees, (e) 40 trees, (f) 50 trees.

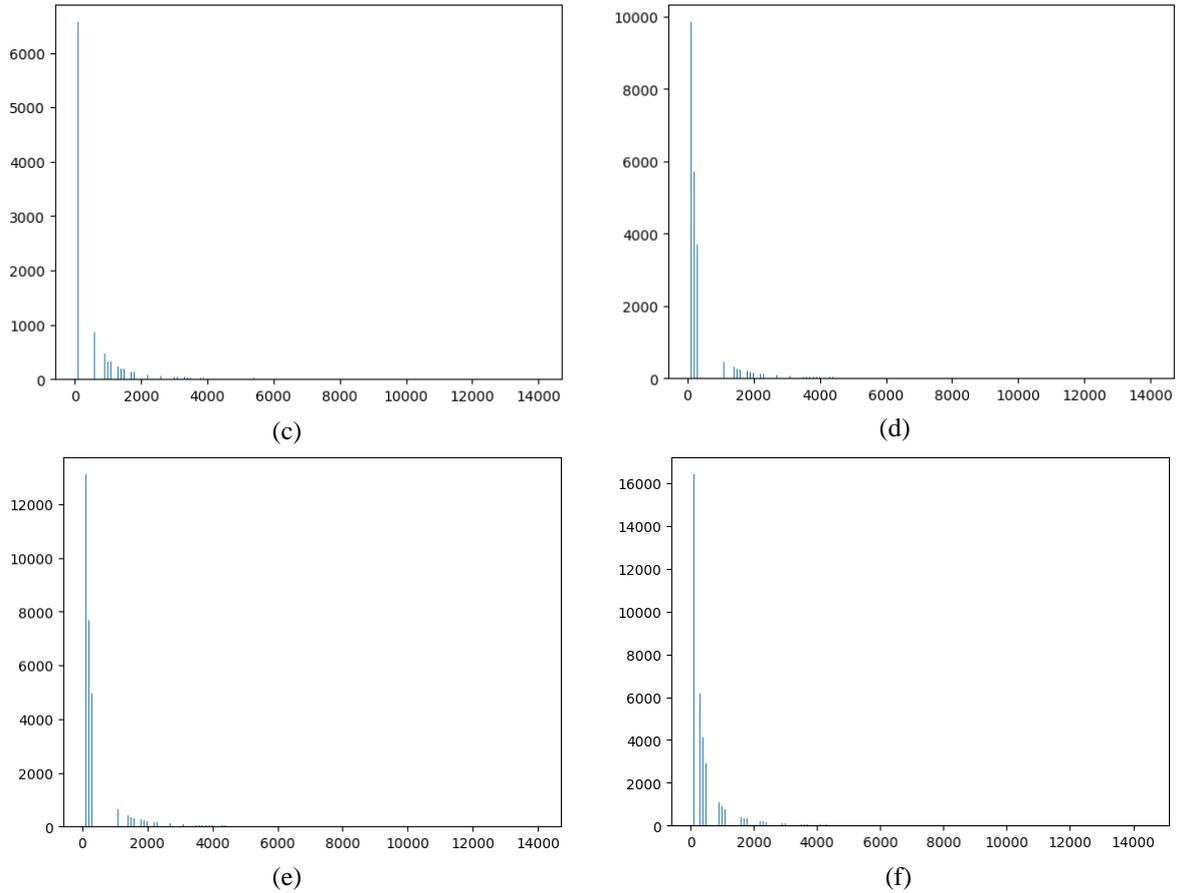

**Fig. 9**. The impact factor of branches for data with 10 clusters in each feature for random forest with (a) one tree, (b) 10 trees, (c) 20 trees, (d) 30 trees, (e) 40 trees, (f) 50 trees.

## 1. Conclusion

We presented a new approach to the RF learning method. The main idea is to bound the depth of trees in the RF method to reduce the size of the trees. This helps us to construct RFs with more trees with fewer depths. In the other words, this would allow us to have more views through more trees with less complexity.

The results have shown that having more informative features in a dataset leads to the need for more trees in the corresponding RF to reach maximum accuracy. In the standard RF, having trees with maximum capable depth not only increases the complexity but also does not lead to a better performance. The results have shown that the maximum depth of trees is not needed to reach the highest accuracy in a given problem. In other words, the trees can be bounded and consequently make the trees smaller. The above two findings suggest that RFs can be constructed with higher numbers of smaller trees to reach better accuracy and performance.

Experimental results on a standard dataset and our Down Syndrome datasets have shown that our IVRD algorithm is more accurate compared to the standard RF. In addition, testing the effect of the increase in the number of RF informative features in dataset has indicated that having features in deeper part of the trees have lees effect on the performance of RF. It also showed that for each number of informative features there is an effective depth that IVRD could find and bound the trees to it. Additionally, it showed that the sample size of a dataset does not have any effect on this trend and the result of IVRD.

The future work would be to use the proposed method on more datasets and applications, such as document retrieval and handwriting recognition. Furthermore, we would further try to provide more accurate time and space complexities analysis. Also, we would experimentally test the impact of reducing the depth on the time and space complexities of the generated RFs. The study of the impact of the noise on the proposed approach would be studied later, too.

**Acknowledgment**